\documentclass{article}
\usepackage{ijcai16}

\usepackage{times}

\usepackage{graphicx}
\usepackage{subfigure}          
\usepackage{amsmath}
\usepackage{amssymb}

\title{Input Aggregated Network for Face Video Representation}
\author{Zhen Dong\textsuperscript{1}, Su Jia\textsuperscript{2}, Chi Zhang\textsuperscript{1}, Mingtao Pei\textsuperscript{1} \\
1. Beijing Laboratory of IIT, School of Computer Science, Beijing Institute of Technology, Beijing, China\\
2. Department of Applied Mathematics and Statistics, Stony Brook University, Stony Brook, USA \\}
\begin{document}

\maketitle

\begin{abstract}
  Recently, deep neural network has shown promising performance in face image recognition.
  The inputs of most networks are face images, and there is hardly any work reported in literature on network with face videos as input.
  To sufficiently discover the useful information contained in face videos, we present a novel network architecture called input aggregated network which is able to learn fixed-length representations for variable-length face videos.
  To accomplish this goal, an aggregation unit is designed to model a face video with various frames as a point on a Riemannian manifold, and the mapping unit aims at mapping the point into high-dimensional space where face videos belonging to the same subject are close-by and others are distant.
  These two units together with the frame representation unit build an end-to-end learning system which can learn representations of face videos for the specific tasks.
  Experiments on two public face video datasets demonstrate the effectiveness of the proposed network.
\end{abstract}

\section{Introduction}
\begin{figure}
    \centering
    \subfigure[]{
        \label{fig:covvsnet-a}
        \includegraphics[width=0.45\textwidth]{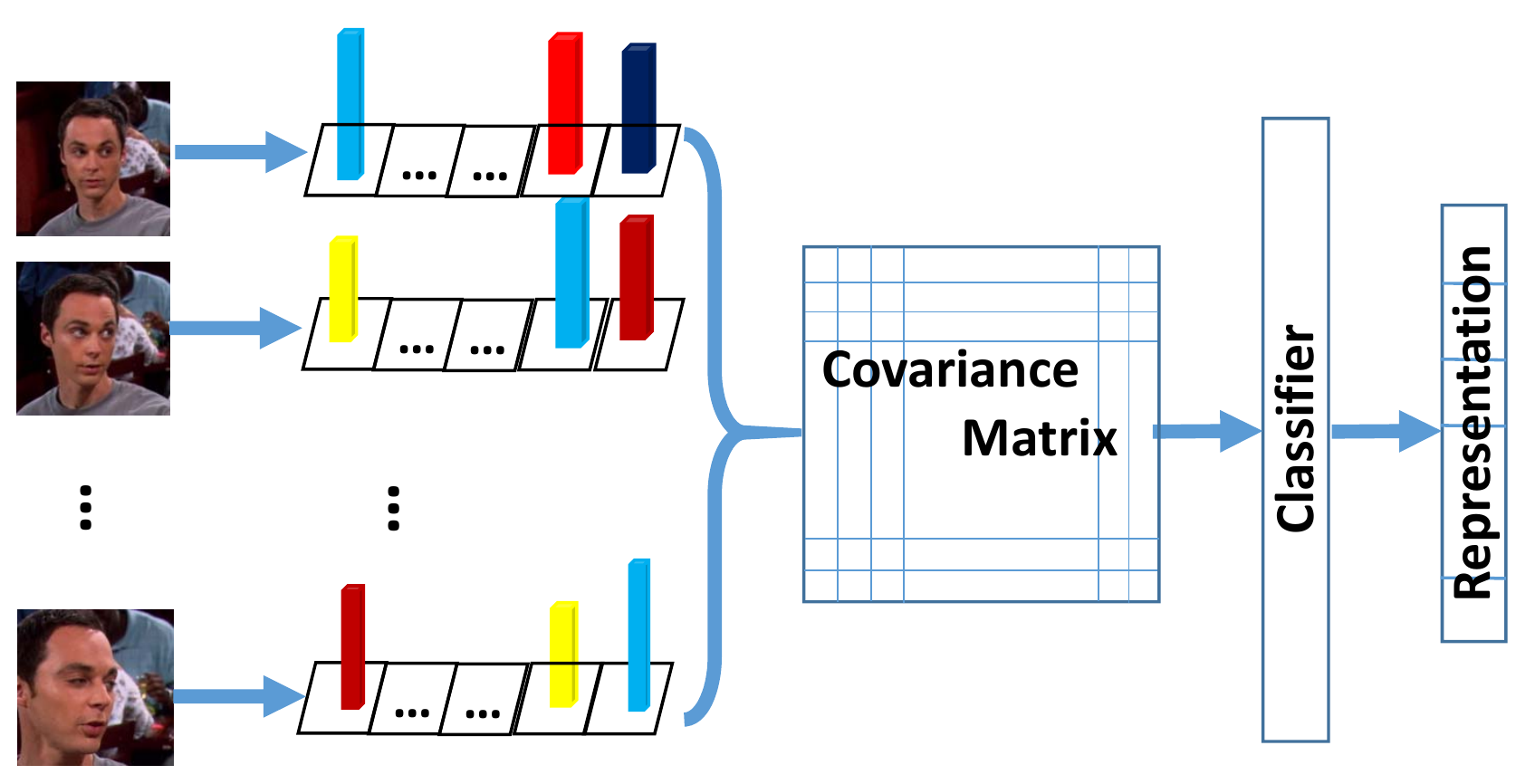}}
    \subfigure[]{
        \label{fig:covvsnet-b}
        \includegraphics[width=0.45\textwidth]{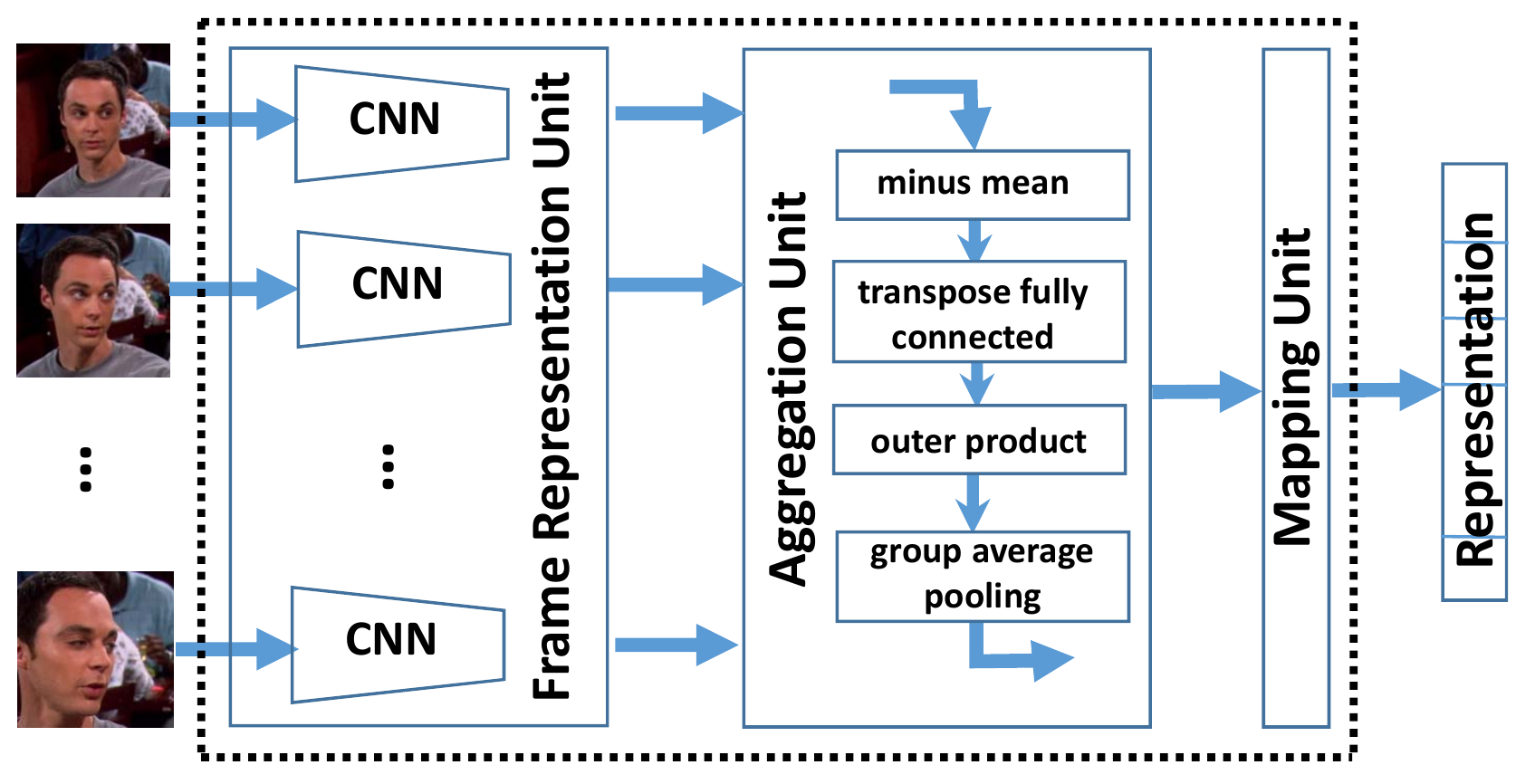}}
    \caption{The illustrations of the traditional face video recognition method and our method. The traditional method (a) has three uncorrelated steps: extracting frame features, modeling face video clip, and learning classifier. Only the classifier learning procedure treats the final recognition task as optimal principle. Differently, the input aggregated network (b) integrates frame representation unit, aggregation unit, and mapping unit into an end-to-end system to learn the mapping from face videos to representations, and all the units serve for the final task.}
    \label{fig:covvsnet}
\end{figure}

Video-based face recognition under uncontrolled environments is a challenging task due to large intra-class variations in pose, lighting, and facial expressions.
There are lots of video-based face recognition methods to deal with these problems \cite{wang2015discriminant,huang2015face,hamm2008grassmann,shakhnarovich2002face}. These methods often contain three steps: extracting frame features, modeling face video clip, and learning classifier.
The three steps are independently conducted and might not be optimally compatible, \emph{i.e.}, the frame feature and the model of a face video might not be discriminative enough for the final recognition task.
Take, for example, the covariance descriptor \cite{huang2015face,li2015hierarchical}, a usual method of face video representation.
The covariance features involve two procedures: extracting low-level features for each frame separately, and fusing these features by computing a covariance matrix, as shown in Figure~\ref{fig:covvsnet-a}.
Both of the procedures are designed independently without taking the final task into account, which will degrade the performance.

As a quite effective tool for building an end-to-end learning system, deep learning has shown promising performances in face recognition.
The representatives are Deep Face \cite{taigman2014deepface,taigman2015web}, DeepID series \cite{sun2013hybrid,sun2014deep,sun2015deeply}, and FaceNet \cite{schroff2015facenet}.
However, all these works use face images as input, and there is hardly any work reported in literature dedicated to design network with face videos as input.
To take full advantage of the useful information contained in videos, we propose the input aggregated network to learn face video representations.

The input aggregated network inputs are variable-length face videos, and outputs are fixed-length representations of the videos.
As shown in Figure \ref{fig:covvsnet-b}, the proposed network contains three units: frame representation unit, aggregation unit, and mapping unit.
In the frame representation unit, a deep CNN is used to learn representation of each frame.
Previous studies \cite{krizhevsky2012imagenet,oquab2014learning} show that the CNN features perform better than many hand-crafted features, so the frame representation unit is able to provide good features for the aggregation unit.
The aggregation unit aims at modeling several frame features of a face video as one Riemannian manifold point.
To achieve this goal, four layers are involved in this unit: minus mean layer, transpose fully connected layer, outer product layer, and group average pooling layer.
With these four layers, we will show that the symmetric positive definite manifold \cite{arsigny2007geometric} and Grasmann manifold \cite{hamm2008grassmann} can be characterized by the aggregation unit.
The mapping unit serves as the map from the low-dimensional Riemannian manifold to high-dimensional space where similar face videos are close by and dissimilar faces are far away.
Considering the map is usually reversible, a deep auto-encoder is expected to well approximate the map owing to its powerful ability on describing highly complex non-linear function.
These three units are optimally combined in the input aggregated network to build an end-to-end learning system where each unit is compatible with others for the final task.

To evaluate the effectiveness of the proposed network, we do experiments of face identification on two datasets: IJB-A \cite{klare2015pushing} and VIPL-TV \cite{li2015face}.
On both dataset, the open-set problem of face identification is considered, \emph{i.e.}, the probe image may not in the gallery set.
In the IJB-A dataset, this is accomplished by removing some videos of specific subjects from the gallery set, and the VIPL-TV dataset has an "unknown" class which contains all face videos belonging to none of any other classes.
Results on these two datasets show better performance of the input aggregated network than the state-of-the-art methods.
Furthermore, the proposed network is able to learn face video representations for other various video-based face related tasks, such as face verification, face video retrieval, by designing loss functions for the specific tasks.

The contributions of the paper are two-folds: (1) The input aggregated network is able to generate fixed-length representations for variable-length face videos. (2) Based on the input aggregated network, a face video identification method is proposed and achieves the comparable performances with other state-of-the-art methods.

The remainder of this paper is organized as follows:
Sec.\ref{sec:related} reviews the related work, including face video representation methods and deep learning on face recognition.
In Sec.\ref{sec:proposed}, we elaborate the architecture and learning policy of the proposed input aggregated network.
Sec.\ref{sec:general} shows that the symmetric positive definite and Grassmann manifolds can be involved in the framework of the proposed network.
The experiments and discussions are presented in Sec.\ref{sec:experiments}, and Sec.\ref{sec:conclusions} concludes this paper.

\section{Related Works}
\label{sec:related}

\subsection{Face Video Representation}
\label{sec:related-vfr}

Face video representation has been widely studied in recent years, and lots of works are proposed.
Most of them treat the face video as a set of frame images, \emph{i.e.} image set based classification.
Closely-related to representation, another issue is the metric definition between face video representations.
One prevalent method represents face image set by a Guassian distribution \cite{arandjelovic2005face,shakhnarovich2002face}, and the Kullback-Leibler divergence is used to measure the similarity between image sets.
The estimated model by this kind of method will not properly fit the real word data distribution when real world data doesn't follow the Gaussian assumption.
Other methods use subspace learning technology to describe face image sets, including linear subspace and manifold.
The principle angle between subspaces is often used as the metric in linear subspace based methods, such as Mutual Subspace Method \cite{yamaguchi1998face}, Orthogonal Subspace Method \cite{oja1983subspace}, and their variants \cite{kim2007discriminative,li2009boosting}.
The linear subspace methods may not work well when the face sets contain complex data variations.
Instead, some methods assume that the face images lie on a nonlinear manifold and propose several manifold distance metrics, such as manifold-manifold distance \cite{wang2012manifold}, manifold distance analysis~\cite{wang2009manifold}, covariance discriminative learning~\cite{wang2012covariance}.
These methods treat frame feature extraction and set modeling as independent procedures, which is different from the input aggregated network.
The input aggregated network builds an end-to-end learning system which involves these two procedures to learn good representations for face videos.

\subsection{Deep Learning on Face Recognition}
\label{sed:related-dlface}

Recently, deep learning has shown promising performances on face recognition.
\citeauthor{taigman2014deepface} \shortcite{taigman2014deepface,taigman2015web} proposed the ``DeepFace" which addresses the face verification problem by a nine-layer deep neural network with $120$ million parameters, and reduced the error of the state-of-the-art by more than $25\%$ on the LFW dataset.
\citeauthor{sun2015deeply} \shortcite{sun2015deeply,sun2015deepid3} proposed the ``DeepID" series which achieve exciting results on both LFW and YouTube Faces datasets for both face identification and verification.
\citeauthor{schroff2015facenet} \shortcite{schroff2015facenet} trained a deep CNN called ``FaceNet" with nearly $200$M face images of about $8$M subjects for face recognition and clustering.
The FaceNet maps face image to a compact representation in Euclidean space and achieves a new record accuracy on both LFW and YouTube Faces datasets.
\citeauthor{chen2015end} \shortcite{chen2015end} aims at building an end-to-end system involving face detection, face association, landmark detection, representation components for face verification with deep convolutional neural network.
Although these works have conducted experiments on face video datsets, such as YouTube Faces, and IJB-A datasets, the networks in these works still take face frame image as input and the relationship between frames is ignored.
\citeauthor{lu2015multi} \shortcite{lu2015multi} modeled face image set as manifold, and proposed a multi-manifold deep learning method to learn the maps from face set space to a shared feature subspace.
The distance between different set can be learned, but the representation of face video cannot be learned.
Different from above networks, our network aims at learning a fixed-length representation of variable-length face video for various face-related tasks, such as identification, verification, and retrieval.

\section{Input Aggregated Network}
\label{sec:proposed}
To represent a face video, three steps are required: representing each face frame, modeling the video clip, and mapping the video representation for the specific task.
Corresponding to the three steps, the input aggregated network contains three units: frame representation unit, aggregation unit, and mapping unit.

\subsection{Frame Representation Unit}
\label{sec:proposed-fru}
As described in Sec.\ref{sed:related-dlface}, deep CNN has shown impressive performance on face image representation in recent years, and previous studies \cite{krizhevsky2012imagenet,oquab2014learning} show that the CNN features perform better than many hand-crafted features.
Motivated by these work, a deep CNN is used as for learning features of each frame in the face video.
As a representative of deep CNN in face recognition, the DeepID~\cite{sun2014deep} significantly promoted performances than previous methods.
The DeepID contains convolutional layers, fully connected layers, ReLU activation layers, and max-pooling layers.
The ReLU activation layers are behind convolutional layers or fully connected layers.
For simplicity, we use $\textbf{L}_{1-4}$ to represent the $4$ convolutional layers, and $\textbf{L}_{5,6}$ describe the $2$ linear layers.
The outputs of $\textbf{L}_{5}$ are features with the dimension of $160$.
The $\textbf{L}_{6}$ is followed by a softmax classifier to generate probability distribution for classification, and the dimension of features in $\textbf{L}_{6}$ is the same with the number of class.

In the frame representation unit, we use the similar architecture as DeepID and make three main modifications.
First, we change the input size from $39 \times 31$ to $48 \times 48$ for more information from larger image.
Second, serving as the feature extractor, a fully connected layer $\textbf{L}_{5.5}$ with $32$ dimension is added between $\textbf{L}_{5}$ and $\textbf{L}_{6}$ to reduce the computation burden of the next two stages.
Third, the convolutional layers $\textbf{L}_{1}$, $\textbf{L}_{2}$ and fully connected layers $\textbf{L}_{5}$, $\textbf{L}_{5.5}$ before ReLU layers are initialized as \cite{he2015deep} to accelerate the convergence.
The deep CNN is pre-trained on the large-scale face image dataset CASIA-WebFace \cite{yi2014learning} which has $494,414$ face images of $10,575$ subjects to obtain good initializations for face representation.
Please note that the used DeepID-like network has a relatively simple architecture for fast computation, and other deeper network with higher performance such as VGG-Net \cite{simonyan2014very} and GoogLeNet \cite{szegedy2014going} can also be used in this unit instead of the DeepID-like architecture.

\subsection{Aggregation Unit}
\label{sec:proposed-au}
\begin{figure}[t]
\begin{center}
\includegraphics[width=0.45\textwidth]{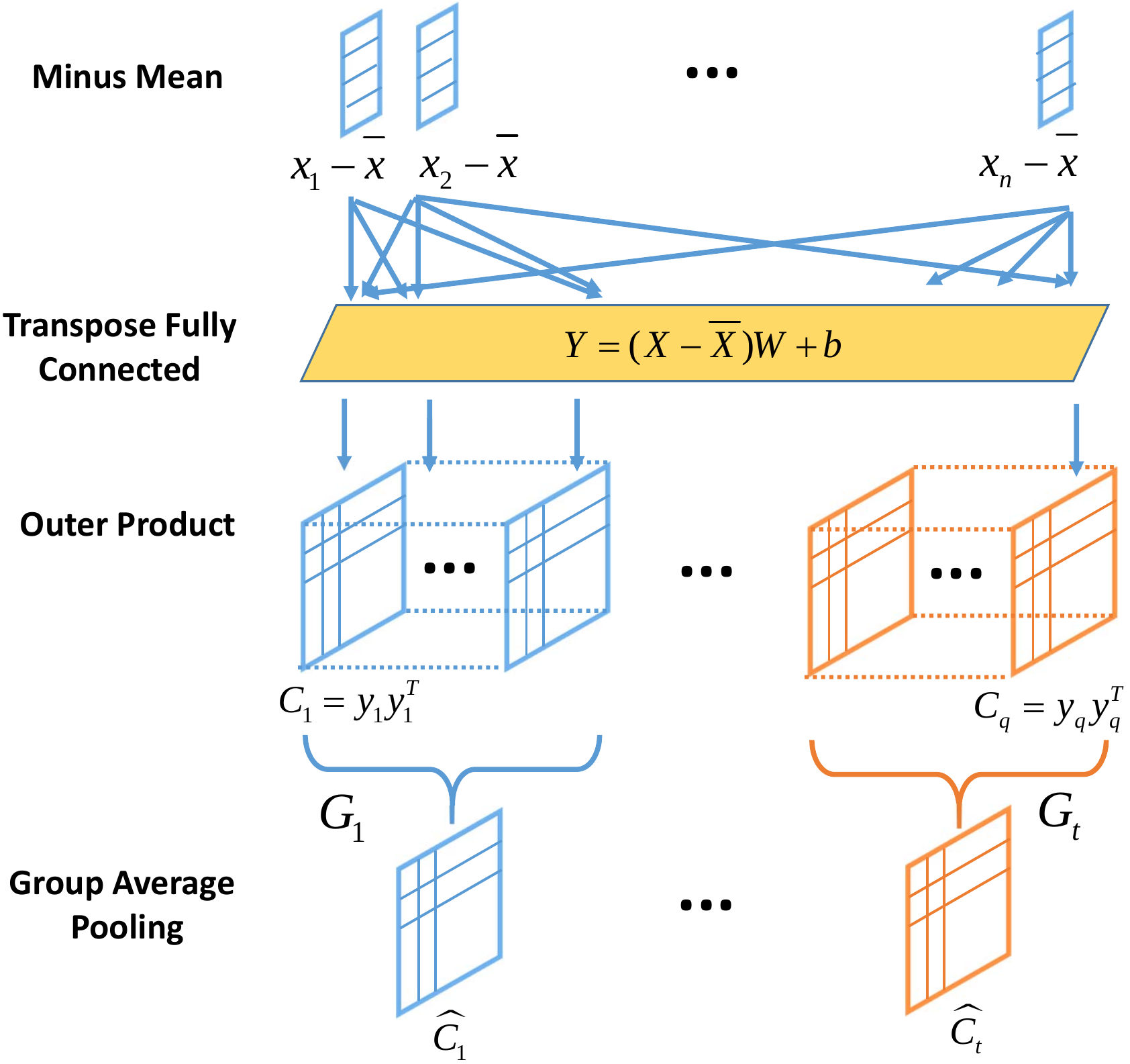}
\end{center}
\caption{The architecture of the aggregation unit. The aggregation unit contains four layers: minus mean layer, transpose fully connected layer, outer product layer, and group average pooling layer.}
\label{fig:aggreunit}
\end{figure}

The aggregation unit aims at modeling the variable-length frame representations as fixed-length Riemannian manifold points.
The architecture of the aggregation unit is shown in Figure \ref{fig:aggreunit}.
The aggregation unit contains four layers: minus mean layer, fully connected layer, outer product layer, and group average pooling layer.

Let $\textbf{X}=[\textbf{x}_{1}, \textbf{x}_{2}, ..., \textbf{x}_{n}] \in \mathbb{R}^{d \times n}$ be the CNN features of the $n$ frames in a face video from the frame representation unit.
The mean vector of these features are calculated as
\begin{eqnarray}
\begin{aligned}
\label{eq:mean}
     \overline{\textbf{x}} = \frac{1}{n}\sum_{i=1}^{n}\textbf{x}_{i}.
\end{aligned}
\end{eqnarray}
The \textbf{Minus Mean Layer} makes the features from frame representation unit to be zero-mean by subtracting the mean vector from all the features as $[\textbf{x}_{1}-\overline{\textbf{x}}, \textbf{x}_{2}-\overline{\textbf{x}}, ..., \textbf{x}_{n}-\overline{\textbf{x}}]$.
The followed \textbf{Transpose Fully Connected Layer} is similar to fully connected layer which provides a linear transformation to its inputs, the difference is that the transpose fully connected layer conducts the linear transformation on the transpose of its input matrix, which aims at finding the optimal linear combination of its inputs to form the subspace of the input mean-subtracted features:
\begin{eqnarray}
\begin{aligned}
\label{eq:fclayer}
     \textbf{Y} = (\textbf{X}-\overline{\textbf{X}})\textbf{W}+\textbf{b},
\end{aligned}
\end{eqnarray}
where $\textbf{W} \in \mathbb{R}^{n \times q}$ and $\textbf{b} \in \mathbb{R}^{d \times q}$ are the parameters of the transpose fully connected layer, and $\overline{\textbf{X}} \in \mathbb{R}^{d \times n}$ is mean matrix whose columns are all $\overline{\textbf{x}}$.
The transpose fully connected layer is able to characterize the respective importance of the frame features in the video by the corresponding coefficients, and can be implemented by a convolutional layer with kernel size of $1 \times 1$.
For each transformed feature $\textbf{y}_{i}$, the \textbf{Outer Product Layer} computes the outer product of $\textbf{y}_{i}$ and itself:
\begin{eqnarray}
\begin{aligned}
\label{eq:oplayer}
     \textbf{C}_{i} = \textbf{y}_{i} \otimes \textbf{y}_{i} = \textbf{y}_{i}\textbf{y}_{i}^{\top}, \ \  i =1,2,...,n,
\end{aligned}
\end{eqnarray}
where $\textbf{C}_{i} \in \mathbb{R}^{d \times d}$ is a symmetric positive semi-definite matrix.
The \textbf{Group Average Pooling Layer} converts the variable-lengths feature into fixed-length features by partition $\textbf{C}_{i}(i=1,2,...,n)$ into $t$ groups according to its frame order, and operating average pooling in each group along with the frame dimension.
\begin{eqnarray}
\begin{aligned}
\label{eq:gaplayer}
     \widehat{\textbf{C}}_{i} = \frac{1}{|\mathcal{G}_{i}|} \sum_{\textbf{C}_{j} \in \mathcal{G}_{i}} \textbf{C}_{j}, \ \  i =1,2,...,t,
\end{aligned}
\end{eqnarray}
where $\mathcal{G}_{i}$ is the $i$-th group of the matrices.
Since there are $n$ matrices coming from the outer product layer, each group in the group pooling layer process $|\mathcal{G}_{i}| = n/t$ matrices.
The group are consecutive in the frame dimensionality, so the aggregation unit is able to hold temporal information for face representation in a certain degree.
With these four layers, the aggregation unit maps $n$ CNN frame representations as input to $t$ matrices as features, where $n$ in the length of variable-length video and $t$ is the number of the fixed-length features.
The $t$ matrices are used as the input of the followed mapping unit.
Since there are four parameters $\textbf{W}$, $\textbf{b}$, $q$, and $t$, an aggregation unit can be denoted as $\mathcal{A}=\{\textbf{W}, \textbf{b}, q, t\}$.

The overall input aggregated network is expected to be an end-to-end system.
Since the aggregation unit forms a directed acyclic graph, it can be trained by back propagation algorithm which propagates the gradients of the loss function to the four layers.
For $i$-th group of the group average pooling layer, let the gradient respect to $\widehat{\textbf{C}}_{i}$ be $\partial l / \partial \widehat{\textbf{C}}_{i}$, the gradient respect to $\textbf{y}_{j}$ is
\begin{eqnarray}
\begin{aligned}
\label{eq:bp-aggre}
     \frac{\partial l}{\partial \textbf{y}_{j}} = \frac{1}{|\mathcal{G}_{i}|} \Bigl[ & (\frac{\partial l}{\partial \widehat{\textbf{C}}_{i}})^{\top} + \frac{\partial l}{\partial \widehat{\textbf{C}}_{i}} \Bigl] \textbf{y}_{j}, \\
     & \ \ \textbf{C}_{j} \in \mathcal{G}_{i}, \ \  i =1,2,...,t.
\end{aligned}
\end{eqnarray}
Eq.(\ref{eq:bp-aggre}) gives the gradients of outer product and group average pooling layers.
Since the transpose fully connected layer is implemented by the convolutional operation with kernel size of $1 \times 1$, the gradient of this layer is easy to obtain similar to the convolutional layer.
The minus mean layer holds gradient unchanged for the reason that only a constant vector is subtracted in this layer.
With the computed gradients, the aggregation unit is able to be embedded into the input aggregated network for end-to-end training.

The matrices from the outer product layer are symmetric positive semi-definite, and the average pooling in each group holds this property.
The outputs of the group average pooling are even symmetric positive definite matrices when $n/t \gg d$.
As proved in \cite{arsigny2007geometric}, the space of symmetric positive definite matrices form a Lie group which is a Riemannian manifold.
Therefore, the output $t$ matrices are all lie on a Riemannian manifold.
In Sec.\ref{sec:general}, we will show that several usually used manifolds in face video representation can be involved in the framework of the aggregation unit.

\subsection{Mapping Unit}
\label{sec:proposed-mu}
Obtaining the Riemannian manifold points which represent face videos, previous methods usually map these data points into a reproducing kernel Hilbert space by kernel methods.
The mapping unit in the input aggregated network aims to learn the map by the deep network with multiple layers.
Let $\mathcal{R}$ be the Riemannian manifold, and $\mathbf{S}$ be the high-dimensional space, the goal of the this unit is learning the map
\begin{eqnarray}
\begin{aligned}
\label{eq:mapdef}
     \textbf{F}:\mathcal{R}\rightarrow\mathbf{S}.
\end{aligned}
\end{eqnarray}
The mapping function is highly complex and nonlinear in general, so we use a deep neural network with $h$ stacked hidden layers to model the mapping.
Let $f_{i}$ be the mapping function of the $i$-th layer, and $\textbf{z}_{i}$ be the output feature representation of the $i$-th layer, we thus have
\begin{eqnarray}
\begin{aligned}
\label{eq:submapdef}
     \textbf{z}_{i}=
     \begin{cases}
      f_{i-1}(\textbf{z}_{i-1})=\tau(\textbf{W}_{i-1}\textbf{z}_{i-1} & +\textbf{b}_{i-1}), \\
       &i= 2,3,...,h+1, \\
      \textbf{C}, &i=1,\\
     \end{cases}
\end{aligned}
\end{eqnarray}
where $\textbf{C} \in \mathcal{R}$ is the input point on the Riemannian manifold, $\textbf{W}_{i-1}$ and $\textbf{b}_{i-1}$ are the parameters of the $(i-1)$-th layer, and $\tau$ represents the non-linear activation function, such as sigmoid, tanh, and ReLU.

\begin{figure}[t]
\begin{center}
\includegraphics[width=0.45\textwidth]{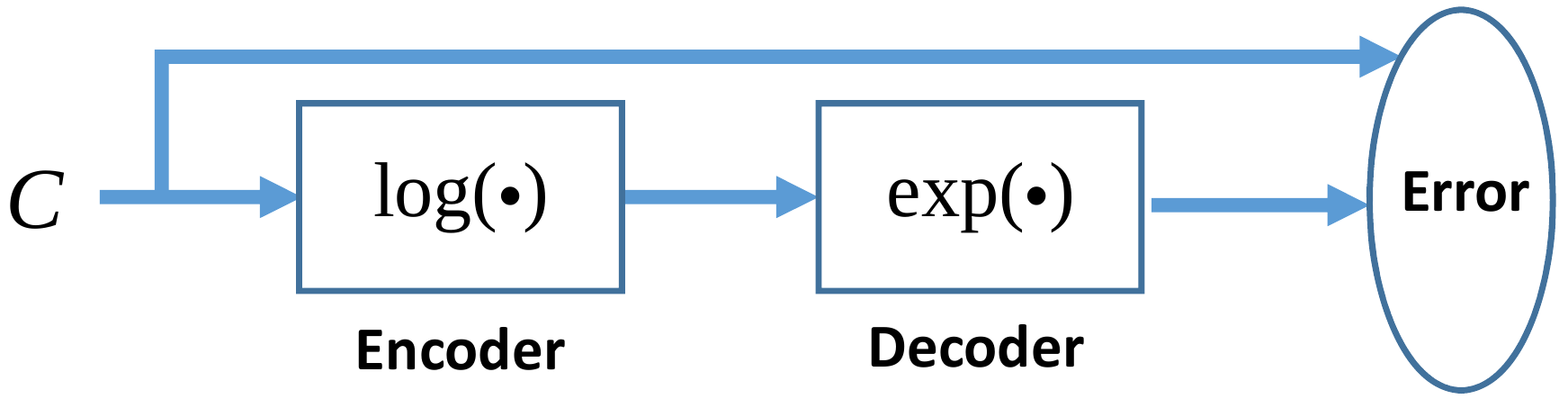}
\end{center}
\caption{The illustration of auto-encoder used for pre-training the mapping unit.}
\label{fig:autoencoder}
\end{figure}

Considering that the map for $\mathcal{R}$ is usually reversible, a deep auto-encoder is trained to obtain a good initializations for the map.
For example, the maps from the symmetrical positive definite Riemannian manifold to its tangent space is $\log(\cdot)$, and the inverse map is $\exp(\cdot)$ as shown in Figure \ref{fig:autoencoder}.
The cost function of the auto-encoder is
\begin{eqnarray}
\begin{aligned}
\label{eq:alobj}
     \min_{\textbf{F},\textbf{F}^{'}} \frac{1}{N} \sum_{j=1}^{N} \| \textbf{C}^{(j)}-\textbf{F}^{'}(\textbf{F}(\textbf{r}^{(j)})) \|_{2}^{2} + \eta \sum_{k}\|\textbf{W}_{k}\|_{F}^{2},
\end{aligned}
\end{eqnarray}
where $\textbf{F}=\{f_{1},f_{2},...,f_{h}\}$ is the encoder, $\textbf{F}^{'}$ is the decoder, $\textbf{C}^{(j)}$ is the $j$-th point in the training set, and $N$ is the size of training set.

\section{Relation to Manifolds}
\label{sec:general}
In this section, we will show that several usually used methods of face video representation can be involved in the framework of the input aggregated network.

\textbf{Symmetric Positive Definite Manifold:}
All the symmetric positive definite matrices form a Riemannian manifold \cite{arsigny2007geometric}.
In computer vision community, the covariance matrix descriptor which is a special case of symmetric positive definite matrix is often used for face video representation owning to its second order statistic property.
Given a set of features $\mathcal{S}$, the covariance matrix descriptor is calculated as
\begin{eqnarray}
\begin{aligned}
\label{eq:cov-def}
     \textbf{C} = \frac{1}{|\mathcal{S}|}\sum_{s\in\mathcal{S}} (s-\mu_{\mathcal{S}})(s-\mu_{\mathcal{S}})^{\top},
\end{aligned}
\end{eqnarray}
where $\mu_{\mathcal{S}}=\frac{1}{|\mathcal{S}|}\sum_{s\in\mathcal{S}}s$ represents the mean vector of $\mathcal{S}$.
The diagonal elements of the matrix characterize the variance of each feature dimension, and the off-diagonal elements encode the respective correlations between feature dimensions.
The distance between two covariance matrices, $\textbf{C}_{a}$ and $\textbf{C}_{b}$ is measured by the Log-Euclidean distance which is a geodesic distance calculated with Euclidean computations in the domain of matrix logarithms:
\begin{eqnarray}
\begin{aligned}
\label{eq:cov-dis}
     d(\textbf{C}_{a},\textbf{C}_{b}) = \|\log(\textbf{C}_{a}) - \log(\textbf{C}_{b}) \|_{F},
\end{aligned}
\end{eqnarray}
where $\|\cdot\|_{F}$ denotes the Frobenius norm of a matrix, and $\log(\cdot)$ as the matrix logarithm operator is computed as
\begin{eqnarray}
\begin{aligned}
\label{eq:logmatrix}
     \log(\textbf{C}) = \textbf{U}\log(\Lambda)\textbf{U}^{\top},
\end{aligned}
\end{eqnarray}
where $\textbf{C} = \textbf{U}\Lambda\textbf{U}^{\top}$ with $\Lambda = diag(\lambda_{1}, \lambda_{2}, ..., \lambda_{d})$ is the eigen decomposition of $\textbf{C}$, and $\log(\Lambda) = diag(\log(\lambda_{1}), $ $\log(\lambda_{2}), ..., \log(\lambda_{d}))$.

\begin{figure}[t]
\begin{center}
\includegraphics[width=0.45\textwidth]{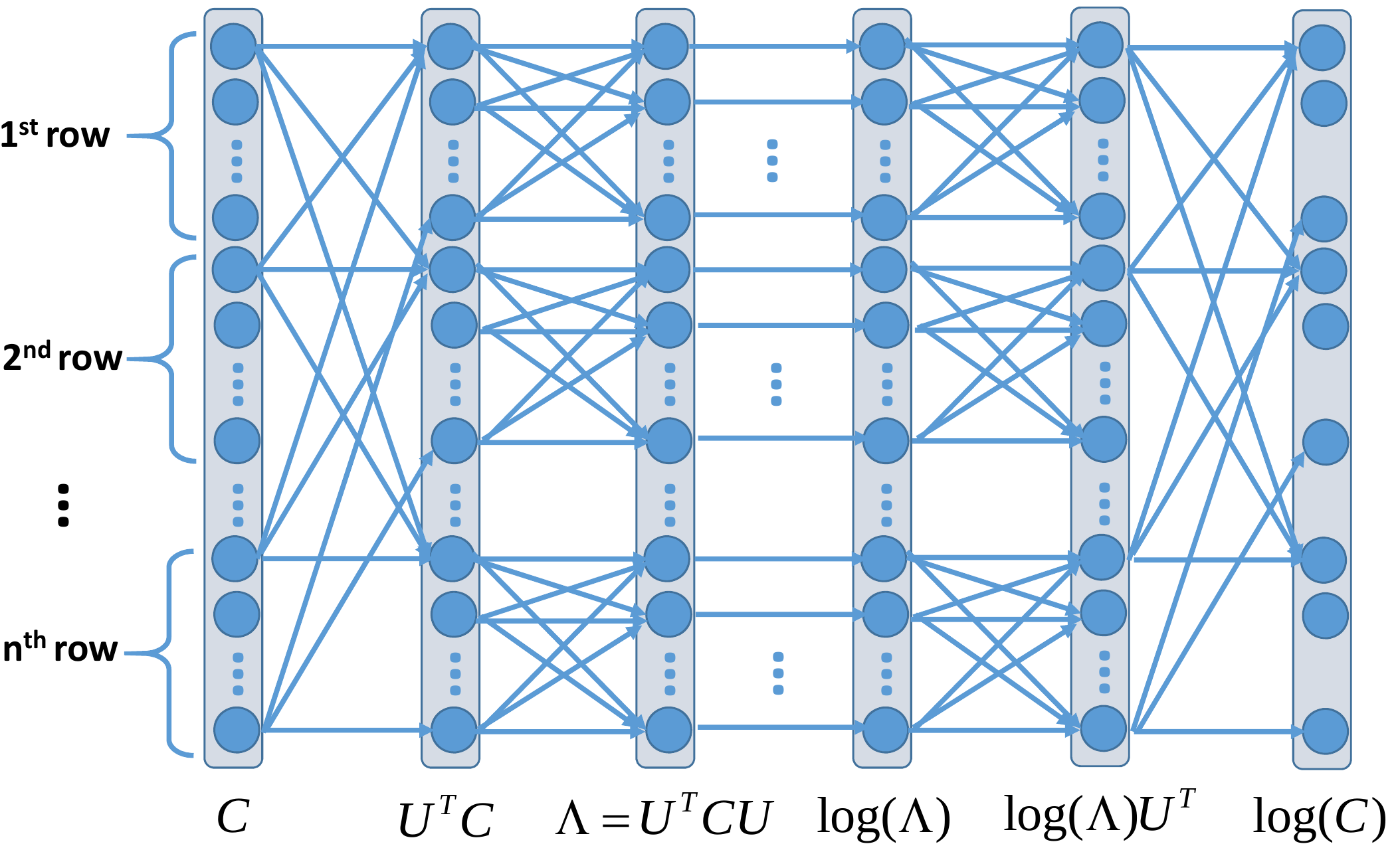}
\end{center}
\caption{The mapping unit is able to represent the matrix logarithm operator $\log(\cdot)$. Please note that not all links are drawn for the sake of simplicity.}
\label{fig:mappingunit}
\end{figure}

The covariance matrix descriptor can be written in the form of aggregation unit by $\mathcal{A}=\{\textbf{I}, \textbf{O}, d, 1\}$ where $\textbf{I} \in \mathbb{R}^{d \times d}$ is the identity matrix, and $\textbf{O} \in \mathbb{R}^{d \times n}$ is a matrix with all elements as $0$.
As shown in Figure \ref{fig:mappingunit}, the non-linear mapping, matrix logarithm operator $\log(\cdot)$, is able to be represented by the mapping unit since $\textbf{U}$ and $\textbf{U}^{\top}$ performs linear transformation which is able to be implemented by the fully connected layer, and the feasibility of approximating the logarithm operation of number by neural network is ensured by the universal approximation theorem \cite{hornik1989multilayer}.

\textbf{Grassmann Manifold:}
Given the feature set $\textbf{X} \in \mathbb{R}^{d \times n}$, the Grassmann manifold represents the feature set by a $q$-dimensional linear subspace which is spanned by an orthonoraml basis matrix $\textbf{Y} \in \mathbb{R}^{d \times q}$:
\begin{eqnarray}
\begin{aligned}
\label{eq:gras-def}
     \textbf{X}\textbf{X}^{\top} \approx \textbf{Y}\Lambda\textbf{Y}^{\top},
\end{aligned}
\end{eqnarray}
where $\Lambda$ and $\textbf{Y}$ are eigenvalue matrix and eigenvector matrix corresponding to the $q$ largest eigenvalues.
The distance between two points on the Grassmann manifold, $\textbf{Y}_{a}$ and $\textbf{Y}_{b}$ is denoted as
\begin{eqnarray}
\begin{aligned}
\label{eq:gras-dis}
     d(\textbf{Y}_{a},\textbf{Y}_{b}) = \frac{1}{\sqrt{2}} \|\textbf{Y}_{a}\textbf{Y}_{a}^{\top} - \textbf{Y}_{b}\textbf{Y}_{b}^{\top} \|_{F}.
\end{aligned}
\end{eqnarray}

The Grassmann manifold can also be represented by the framework of the input aggregated network.
Let $\textbf{W}^{*} \in \mathbb{R}^{n \times q}$ be the transformation matrix from $\textbf{X}$ to $\textbf{Y}$:
\begin{eqnarray}
\begin{aligned}
\label{eq:gras-W-opti}
     \textbf{W}^{*} = \arg \min_{\textbf{W}} \frac{1}{2} \|\textbf{X}\textbf{W} - \textbf{Y} \|_{F}^{2},
\end{aligned}
\end{eqnarray}
which can solve as $\textbf{W}^{*} = \textbf{X}^{\dag}\textbf{Y}$ where $\textbf{X}^{\dag}$ is the pseudo-inverse of $\textbf{X}$, and we have that $\textbf{X}\textbf{W}^{*} \approx \textbf{Y}$.
The $\textbf{b}^{*}$ is set as $\overline{\textbf{X}}\textbf{X}^{\dag}\textbf{Y}$, and the group size is set as $1$.
The output of the aggregation unit is
\begin{eqnarray}
\begin{aligned}
\label{eq:gras-output}
     &\frac{1}{n}
     \big[(\textbf{X}-\overline{\textbf{X}})\textbf{W}^{*}+\textbf{b}^{*}\big]
     \big[(\textbf{X}-\overline{\textbf{X}})\textbf{W}^{*}+\textbf{b}^{*}\big]^{\top}\\
     &=\frac{1}{n}\textbf{X}\textbf{W}^{*}\textbf{W}^{*\top}\textbf{X}^{\top}
     \ = \frac{1}{n}\textbf{Y}\textbf{Y}^{\top},
\end{aligned}
\end{eqnarray}
which is the interchangeably representation of Grassmann manifold with ignoring the constant factor $1/n$.
Therefore, the Grassmann manifold is represented by the aggregation unit as $\mathcal{A}=\{\textbf{X}^{\dag}\textbf{Y},\overline{\textbf{X}}\textbf{X}^{\dag}\textbf{Y},q,1\}$.




\section{Experiments}
\label{sec:experiments}

\subsection{Datasets and Experimental Settings}
We do experiments on two face video datasets: IARPA Janus Benchmark-A (IJB-A) \cite{klare2015pushing}, and VIPL-TV \cite{li2015hierarchical}.

\begin{figure}[t]
\begin{center}
\includegraphics[width=0.45\textwidth]{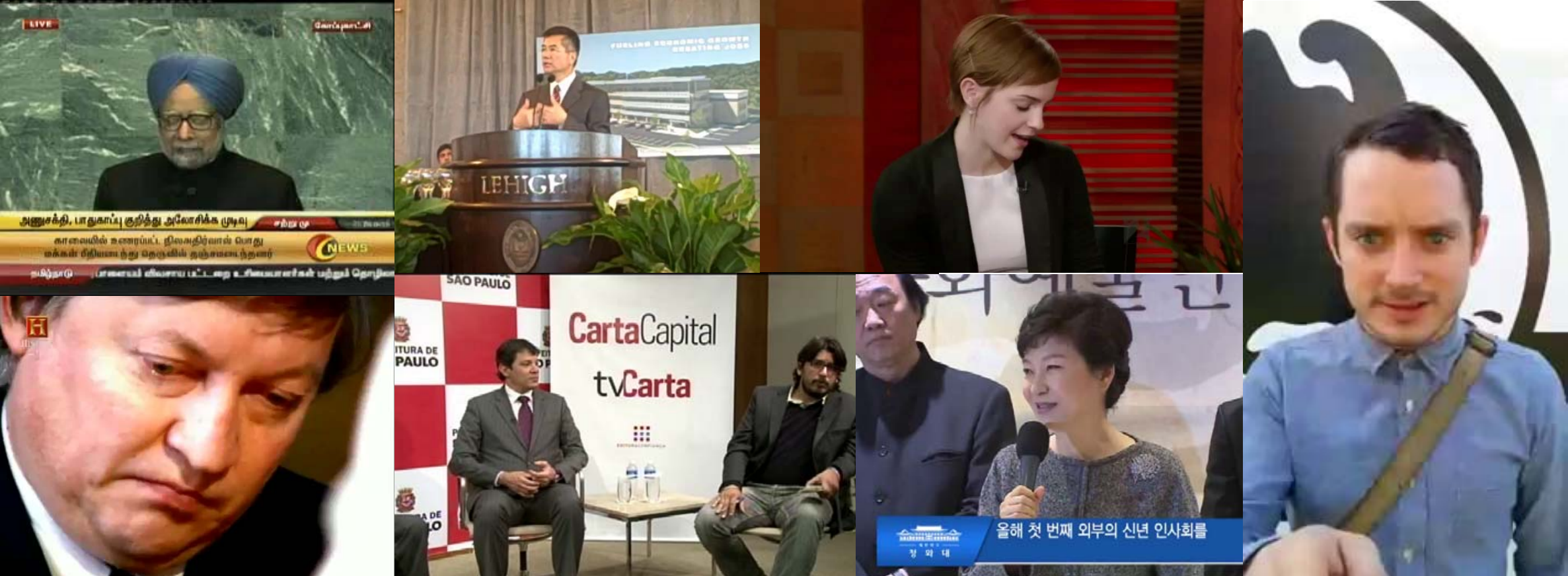}
\end{center}
\caption{The example frames of the IJB-A dataset.}
\label{fig:ijba}
\vspace{-5pt}
\end{figure}

The IJB-A dataset\footnote{This paper makes use of the following data made available by the Intelligence Advanced Research Projects Activity (IARPA): IARPA Janus Benchmark A (IJB-A) data detailed at http://www.nist.gov/itl/iad/ig/facechallenges.cfm. The original videos are contained in the extend version Janus Challenging set 2 (JANUS CS2) which is not publicly available yet.} contains $500$ subjects, and each subject has at least $5$ images and $1$ video to provide a total of $5,392$ images and $2,085$ videos, with an average of $11.4$ images and $4.2$ videos per subject.
The IJB-A dataset doesn't provide the original videos, and the sampled frames from the videos are provided.
The sampled frames are $20,412$ in total.
The example frames of the IJB-A dataset is shown in Figure \ref{fig:ijba}.
The image and videos are all under uncontrolled environments and with challenging factors, such as variations in face orientation, pose, pose, and lighting conditions.

\begin{figure}
    \centering
    \subfigure[]{
        \label{fig:vipltv-bbt}
        \includegraphics[width=0.23\textwidth]{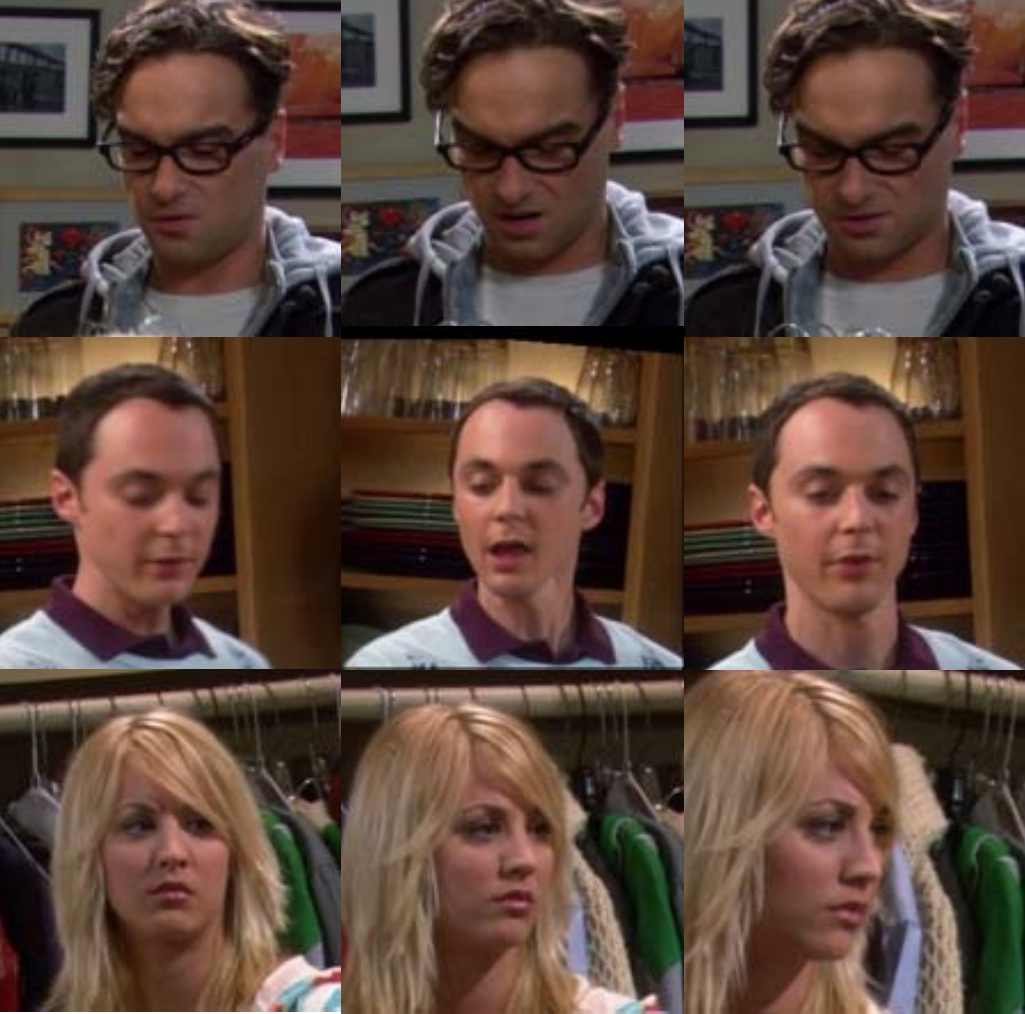}}
    \subfigure[]{
        \label{fig:vipltv-pb}
        \includegraphics[width=0.23\textwidth]{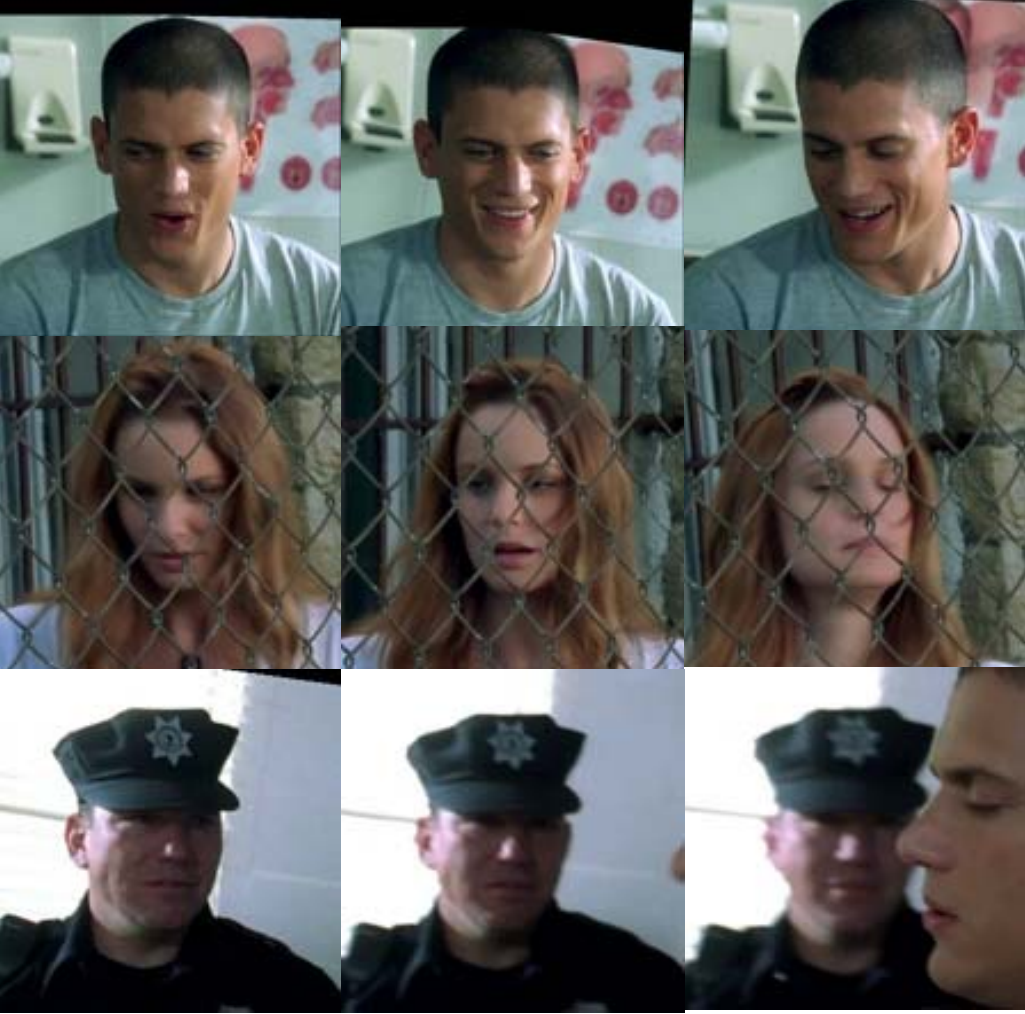}}
    \caption{Example face tracks in VIPL-TV datasets, (a) is from the big bang theory, and (b) is from prison break.}
    \label{fig:vipltv}
\vspace{-5pt}
\end{figure}

The VIPL-TV dataset contains two large scale video collections from two hit American shows, \emph{i.e.}, the Big Bang Theory (BBT) and Prison Break (PB).
These two TV series are quite different in their filming styles.
The BBT is a sitcom with $5$ main characters, and most scenes are taken indoors during each episode of about $20$ minutes long.
Differently, many shots of the PB are taken outside during the episodes with the length of about 42 minutes, which results in a large range of different illumination.
All the face video shots are collected from the whole first season of both TV series, \emph{i.e.}, 17 episodes of BBT, and 22 episodes of PB, and the number of shots of the two sets are $4,667$ and $9,435$, respectively.
Figure \ref{fig:vipltv} shows example face tracks of this dataset.

On both datasets, our network is trained in three stages: pre-training the frame representation unit, pre-training the mapping units, and fine-tuning the whole network.
In all three stages, the base learning rate is set as $0.01$ and reduced by polynomial strategy with power equals to $0.05$.
The momentum and weight decay for optimization are set as $0.9$ and $0.0005$, respectively.
All the codes are implemented by using Caffe deep learning toolbox \cite{jia2014caffe}, and the experiments are conducted on a Titan-X GPU with $12$GB memory.
In the first stage, we pre-train the frame representation unit on the WebFace dataset with the ratio of training sample number to validation sample number as $9:1$.
The batch size is set as $128$, and the total iteration number is set as $5$ million.
In the second stage, the mapping unit is pre-trained with the $\textbf{W}$ and $\textbf{f}$ of the aggregation unit as identity and zero matrices, respectively.
The auto-encoder described in Sec.\ref{sec:proposed-mu} is trained where the batch size is set as $12$ (videos), and the total iteration number is $200$K.
In the third stage, the whole network is fine-tuned with bach size and iteration number of $12$ and $50$K, respectively.
We use two newly published manifold learning methods for face video recognition as our baseline, they are Projection Metric Learning on Grassmann manifold (PML) \cite{huang2015projection} and Discriminative Analysis on Riemannian manifold of Gaussian distributions (DARG) \cite{wang2015discriminant}.
For fair comparisons, the features from the frame representation unit are used to test these two methods.

\subsection{IJB-A Dataset}

The IJB-A evaluation protocol consists of identification over 10 splits.
In each split, $333$ subjects are randomly chosen to be training set, and the remaining $167$ subjects are for testing.
Every testing subject has images or frames randomly sampled into either probe set or gallery set.
The image in gallery set is used as query, and the gallery set represents the imagery contained in an operational database.
To meet the real word applications, the open-set identification protocol is provided, \emph{i.e.}, the subject in probe set may not appear in the gallery set.
The IJB-A dataset achieves this by randomly removing $55$ subjects from the gallery set.

Considering that the IJB-A dataset only contains the sampled frames of face videos and images, and the sampled frames are not consecutive, we treat a face video as a face image set, \emph{i.e.}, the temporal relationship among frames are not taken into account.
Accordingly, the group size in the aggregation unit is set as $1$.
To augment the training data, we form a batch as $16$ frames or images of a subject.
The comparison results are shown in Table.\ref{tab:cr-ijba}, the reported results are the average of ten splits.
The proposed method outperforms other two baseline methods, and the main reason is that the end-to-end training strategy ensures all the three units of the input aggregated network serving for the final identification task.

\begin{table}
\begin{center}
\renewcommand{\arraystretch}{1.0}
\small
\caption{Results on IJB-A Dataset.}
\label{tab:cr-ijba}
\begin{tabular}{lc}
\hline
Methods & Accuracy (\%) \\
\hline
\hline
PML & 30.64 \\
DARG & 34.71 \\
Our Method & 46.25 \\
\hline
\vspace{-20pt}
\end{tabular}
\end{center}
\end{table}

\begin{table}
\begin{center}
\renewcommand{\arraystretch}{1.0}
\small
\caption{Results on VIPL-TV Dataset.}
\label{tab:cr-vipltv}
\begin{tabular}{lcc}
\hline
Methods & Accuracy (BBT) (\%) & Accuracy (PB) (\%) \\
\hline
\hline
PML & 46.20 & 16.47 \\
DARG & 48.59 & 17.11\\
Our Method & 60.32 & 25.81\\
\hline
\vspace{-20pt}
\end{tabular}
\end{center}
\end{table}

\subsection{VIPL-TV Dataset}

Similar to the IJB-A dataset, the open-set identification protocol is used in the experiments of VIPL-TV dataset..
In both BBT and PB of the the dataset, there is an 'unknown' class which consists of subjects that don't belong to any other classes.
We train the network with the 'unknown' subjects as a single class to accomplish the open-set identification.
The experimental results shown in Table.\ref{tab:cr-vipltv} demonstrate the effectiveness of the proposed network.

\section{Conclusions}
\label{sec:conclusions}
A novel network architecture called input aggregated network is proposed for face video representation.
The proposed network contains three units: frame representation unit, aggregation unit, and mapping unit.
These three units form an end-to-end learning system which can learn fixed-length representations for variable-length face videos for specific tasks.
Experiments conducted on two public face video datasets demonstrate the effectiveness of the proposed network in face identification.
The future work is to apply the input aggregated network to other video-based face-related tasks, such as face verification and face video retrieval.


\small
\bibliographystyle{named}
\bibliography{egbib}

\end{document}